\documentclass[letterpaper, 10 pt, journal, table]{IEEEtran} 
\IEEEoverridecommandlockouts                             

\usepackage{graphics} 
\usepackage{float}
\usepackage{cite}
\usepackage{adjustbox}
\usepackage{mathptmx} 
\usepackage{stfloats}
\usepackage{amsmath} 
\usepackage{amssymb}  
\usepackage{todonotes}
\usepackage{multirow}
\usepackage{algorithm}
\usepackage{algpseudocode}
\usepackage{subcaption}
\usepackage{color,soul}
\makeatother
\usepackage[colorlinks, linkcolor=green, citecolor=blue]{hyperref}
\def\BibTeX{{\rm B\kern-.05em{\sc i\kern-.025em b}\kern-.08em
  T\kern-.1667em\lower.7ex\hbox{E}\kern-.125emX}}

\sethlcolor{yellow}
\newcommand{\revision}[1]{#1}
\title{\LARGE \bf
\revision{Action-Conditioned World Model for Goal Plane Guidance in Robotic Neck Ultrasound}
}

\author{ Siqi Fan$^{1,3}$ 
Ran Liu$^{5}$ 
Zixuan Yang$^{3,4}$ 
Xiaoyu Fu$^{4}$ 
Xiaoqing Gao$^{4}$ 
Yingqi Xing$^{5}$ 
Yunhui Liu$^{1}$ 
Hongbin Liu$^{3,4}$
Mingcong Chen$^{2,3}$ 
\thanks{*This work was supported by InnoHK, CUHK T Stone Robotics Institute and the HK RGC AoE under AoE/E-407/24-N.}
\thanks{*This study obtained ethical approval from the Institute of Automation, Chinese Academy of Sciences local ethics committee (study title: Ultrasound Robot Motion Perception and Control Based on Multimodal Sensing, study reference: IA21-2502-020302).}
\thanks{
$^{1}$ Siqi Fan and Yun-hui Liu are with the Department of Mechanical and Automation Engineering, Chinese University of Hong Kong, Hong Kong SAR, 
$^{2}$ Mingcong Chen is with the Department of Biomedical Engineering, City University of Hong Kong, Hong Kong SAR
$^{3}$ Mingcong Chen, Siqi Fan, Zixuan Yang and Hongbin Liu are with the Centre for Artificial Intelligence and Robotics Hong Kong Institute of Science \& Innovation, Chinese Academy of Sciences, Hong Kong SAR, 
$^{4}$ Zixuan Yang, Xiaoyu Fu, Xiaoqing Gao and Hongbin Liu are with the Institute of Automation, Chinese Academy of Sciences, China, 
$^{5}$ Ran Liu and Yingqi Xing are with the Department of Vascular Ultrasound, Xuanwu Hospital, the Capital Medical University.}
\thanks{Correspondence: Mingcong Chen \tt{mingcong.chen@cair-cas.org.hk}

}}

\begin{document}

\maketitle
\thispagestyle{empty}
\pagestyle{empty}

\begin{abstract}
\revision{Ultrasound examinations typically follow specific scanning protocols that require navigating through an ordered sequence of anatomical landmarks and target imaging planes. We present a local action-conditioned world model framework for goal plane probe guidance in robotic ultrasound. }
Autonomous ultrasound tasks often require large numbers of probe-motion trajectories for training, but collecting high-quality demonstrations is labor-intensive and explicit simulators are difficult to build because ultrasound appearance depends on contact, tissue deformation, and view-dependent acoustic artifacts.
We address this problem with a two-stage model-based learning pipeline.
First, a latent conditional diffusion world model predicts future ultrasound observations from recent context frames, probe motions and temporal offset.
Second, a goal-conditioned temporal transformer predicts ordered probe motions and is fine-tuned using rewards from the frozen world model.
Experiments on the self-collected dataset show that the world model preserves action-dependent anatomical structure on target-directed scans. In real-world closed loop experiments, the framework achieves success rates of 70.0\% for carotid guidance and 65.0\% for thyroid guidance.
\revision{These results demonstrate the potential of learned ultrasound dynamics for training goal-directed robotic probe navigation within protocol-constrained organ specific tasks.}
\end{abstract}

\begin{IEEEkeywords}
Medical Robots and Systems; AI-Enabled Robotics; Robotic ultrasound; Model-based Control
\end{IEEEkeywords}

\section{Introduction}\label{sec:intro}

Ultrasound imaging is non-radioactive, non-invasive and provides real time imaging at low cost, which makes it widely used in diagnosis and image guided intervention. However, the ultrasound scanning task is highly dependent on the operator, which requires extensive training and experience to acquire high quality images.
This dependency limits reproducibility and becomes increasingly problematic as ultrasound demand grows faster than the sonographer workforce \cite{won2024sound}, while high workload and burnout further aggravate staffing pressure \cite{younan2022burnout}.
Therefore, robotic ultrasound systems aim to improve scan repeatability, reduce operator burden, and extend access to standardized imaging\cite{jiang2023robotic}.

Autonomous robotic ultrasound has been studied through visual servoing, reinforcement learning, task planners, and learned sonography skills\cite{bi2024machine}.
Prior work has trained policies to navigate probes toward standard planes \cite{li2021autonomous}, robotized artery image acquisition \cite{chen2023fully}, learned reward functions from limited demonstrations \cite{jiang2024intelligent}, and built high level robotic ultrasound agents for guideline driven procedures \cite{chen2025uspilot,bi2026scanning}.
\revision{
These methods demonstrate the potential of learning-based robotic ultrasound guidance. However, demonstration-based policies primarily learn from expert action labels, while existing model-based approaches use learned motion representations. Explicitly incorporating predicted action consequences into policy optimization offers an additional learning signal beyond direct action supervision.}

\begin{figure}[!t]
  \centering
  \includegraphics[width=0.75\linewidth]{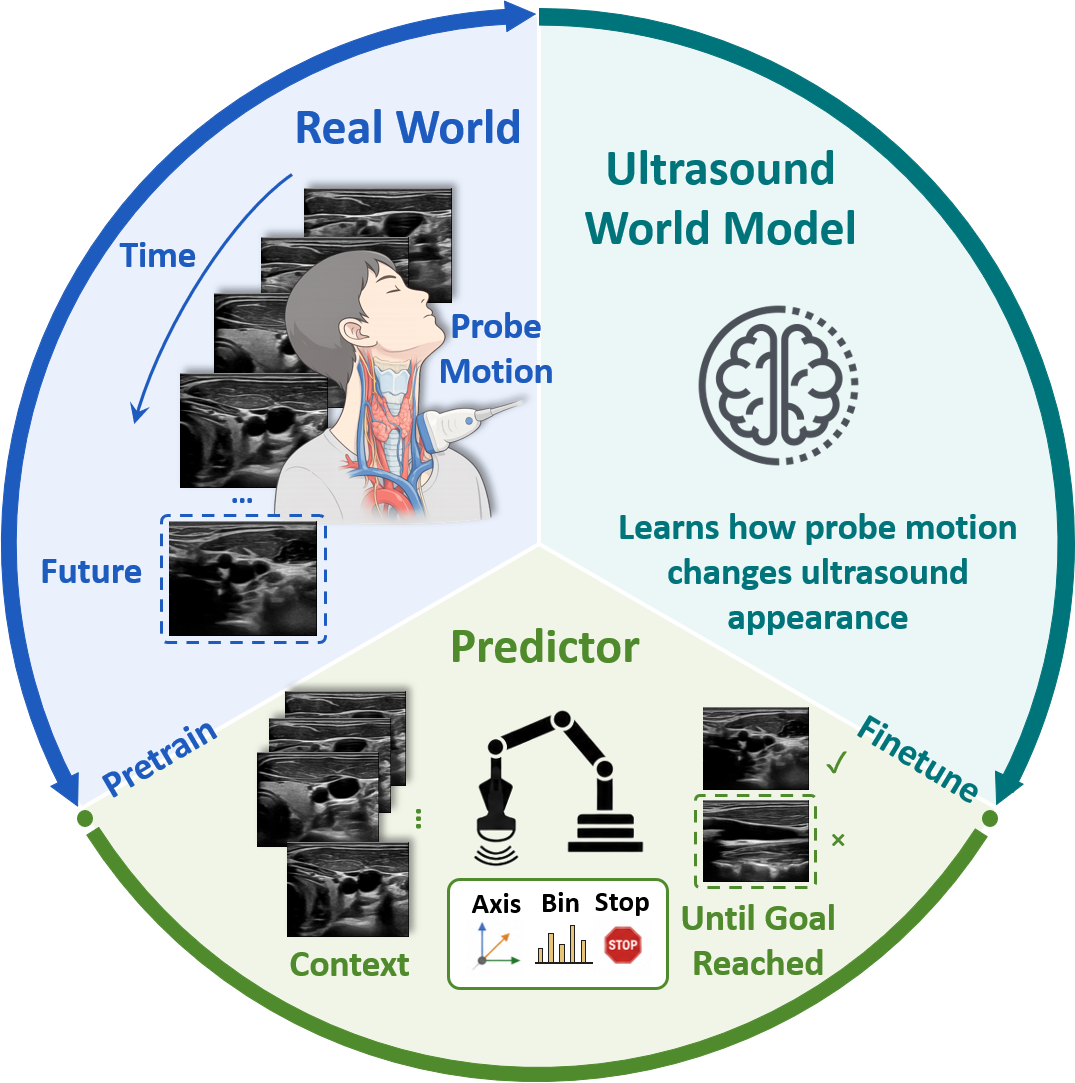}
  \caption{Overview of the action-conditioned world model for goal plane probe guidance in robotic neck ultrasound}
  \label{fig:overview}
  \vspace{-5mm}
\end{figure}

The central challenge is that direct supervision and explicit simulation are both difficult to obtain at scale.
High quality expert trajectories and standard plane annotations are expensive, while physics based ultrasound simulators struggle to capture tissue deformation, acoustic shadowing, probe pressure and subject specific anatomy \cite{ao2026sonogym}.
\revision{
In contrast, synchronized ultrasound images and probe motions can be collected during routine or exploratory scanning without requiring expert-optimal actions or target-plane annotations. This motivates separating ultrasound dynamics learning from goal-conditioned policy learning: a predictive model can learn from a broader range of scanning trajectories, while expert demonstrations and predicted action consequences provide complementary signals for training the guidance policy.
}


\revision{
World models provide a way to connect robot actions with future observations~\mbox{\cite{ebert2018visual}}. In ultrasound guidance, EchoWorld~\mbox{\cite{yue2025echoworld}} employs motion-aware representation pretraining followed by supervised probe-motion regression, while Cardiac Copilot~\mbox{\cite{jiang2024cardiac}} uses a learned spatial model to refine predicted probe motions. Unlike these approaches, our framework retains a separately trained and frozen action-conditioned world model during policy optimization. After supervised initialization, the policy is further optimized using rewards derived from predicted future ultrasound states. This allows the policy to learn from the estimated consequences of its actions rather than relying solely on expert action labels.
Furthermore, the world model is independent of the active scanning goal. The same learned dynamics can be used to evaluate candidate actions against different target planes within the neck region, without retraining the world model. A shared goal-conditioned policy handles the active goal and predicts local corrective motions and STOP decisions for sequential target-plane acquisition.
}

As shown in Fig.~\ref{fig:overview}, our framework contains two stages.
First, a latent conditional diffusion world model is trained from neck ultrasound scanning trajectories to predict future observations from recent context frames, relative probe motion, and temporal offset.
\revision{
Second, a goal-conditioned temporal transformer is initialized through supervised imitation and subsequently fine-tuned with rewards computed from the frozen world model.
}
The reward evaluates whether the predicted future observation moves closer to the active target plane, allowing the policy to refine its actions based on predicted visual progress.
The main contributions of this paper are as follows:
\begin{itemize}
  
    \item \revision{We develop a goal-independent, action-conditioned world model that learns neck ultrasound dynamics without requiring expert-optimal actions or target-plane annotations.}

    \item \revision{We propose a two-stage training framework that uses a frozen world model to refine a goal-conditioned policy through predicted action consequences beyond supervised imitation.}

    \item \revision{We enable sequential goal-plane guidance with a shared policy and STOP prediction, validated through carotid and thyroid robotic ultrasound experiments.}

\end{itemize}





\section{Methodology}

\subsection{Problem formulation}

We consider the problem of protocol-driven navigation during neck ultrasound. At time $t$, the robot observes recent ultrasound frames and chooses a probe motion that moves the image toward the current target standard plane. The challenge is partially observable and action-dependent: the correct motion cannot be determined from the current frame alone, because the usefulness of an action is defined by the future ultrasound observation that it will produce.

We build a self collected image-action dataset on the neck area, $D = \{(I_0,P_0,I_1,P_1,\ldots,I_T,P_T)\}$, where $I_t \in \mathbb{R}^{H\times W\times 3}$ is the ultrasound image at time step $t$ and $P_t$ is the synchronized 6-DoF probe pose in the world coordinate.
For dynamic learning, absolute poses are converted into relative probe motions $a_{i \rightarrow j}$ between selected frames $I_i$ and $I_j$. 
Each target scan is represented as an ordered sequence of standard-plane goals $G_1,\ldots,G_K$, where each goal denotes a target-plane distribution rather than a single image.

\revision{
The two models are trained with different supervision requirements. The world model learns from synchronized ultrasound images and probe motions without requiring target-plane annotations or expert-optimal action labels. In contrast, the goal-conditioned policy is initialized using expert demonstrations with ordered target-plane goals. This separation allows a broader range of neck scanning trajectories to contribute to dynamics learning, while goal-specific supervision is reserved for policy training.}

The first learning problem is action-conditioned ultrasound dynamics. At time $t$ given the latent states of the current $c$ context frames $S_{t-c+1:t}$, a candidate relative action $a_{t \rightarrow t+\Delta t}$, and a temporal offset $\Delta t$, the world model estimates a conditional distribution over the future latent state
\begin{equation}
    W(\hat S_{t+\Delta t} \mid S_{t-c+1:t}, a_{t \rightarrow t+\Delta t}, \Delta t)
\end{equation},
where the latent state $S_t$ can be obtained by an image encoder $enc$ from the ultrasound image $I_t$: $S_t = enc(I_t)$.
This world model predicts the ultrasound observation expected after the probe executes a candidate motion from the current context.

\begin{figure}[t]
  \centering
  \includegraphics[width=0.7\linewidth]{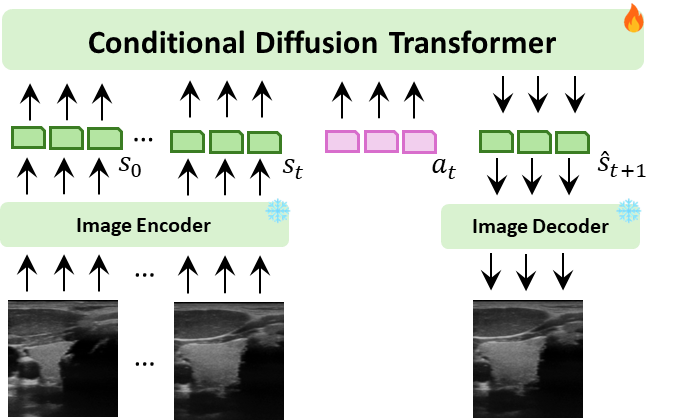}
  \caption{Structure of the action-conditioned latent diffusion ultrasound world model.}
  \label{fig:wm}
  \vspace{-5mm}
\end{figure}

The second learning problem is goal-conditioned action prediction. Given the current latent context $S_{t-c+1:t}$ and the active target-plane representation $G_k$, the policy $\pi$ produces a horizon-length sequence of probe actions $\hat{a}_{t:t+H}$:
\begin{equation}
    \pi(\hat{a}_{t:t+H} \mid S_{t-c+1:t}, G_k)
\end{equation},
where $G_k$ is the $k$ th goal plane in the ordered scanning protocol. The policy is trained with supervised action labels and further guided by the world model. For a predicted action $\hat a_t$, the frozen world model rolls out the corresponding future latent state $\hat S_{t+\Delta t}$, and a reward evaluates whether this predicted state moves toward the active goal plane distribution:
\begin{equation}
    R = r(\hat S_{t+\Delta t}, G_k)
\end{equation},
where $r$ measures both progress toward and proximity to the target-plane distribution. This formulation separates the paper's two central questions: learning how probe motion changes ultrasound appearance, and using that learned action consequence to train a goal directed probe motion policy.

\subsection{Action-conditioned latent diffusion ultrasound world model}
The proposed world model is introduced to make more effective use of large scale ultrasound scanning trajectories. Compared with expert action labels or manually annotated standard planes, image-action-image transitions are much easier to obtain because they are produced continuously during free-hand or robotic scanning. 


As shown in Fig.~\ref{fig:wm}, the model predicts the future ultrasound observation by jointly considering recent visual context, the relative probe motion, and the temporal offset. 
As stated in the problem formulation, we encode ultrasound frames into a compact latent representation using a frozen VAE encoder $enc$. 
Operating in the VAE latent space reduces the spatial redundancy of ultrasound images while preserving the speckle texture, anatomical boundaries, and acoustic shadow patterns that are important for navigation.

To implement the world model $W$, we employ a Conditional Diffusion Transformer (CDiT)~\cite{bar2025navigation} with learnable parameters $\theta$. The function $f_{\theta}$ models the ultrasound dynamics by predicting noise during the diffusion process. The latent dynamics are modeled as follows. 
During training, Gaussian noise $\epsilon$ is added to the target latent according to a diffusion step $\tau$:
\begin{equation}
    S_{t+\Delta t}^{\tau}
    = \sqrt{\bar{\alpha}_{\tau}} S_{t+\Delta t}
    + \sqrt{1-\bar{\alpha}_{\tau}} \epsilon, 
    \quad \epsilon \sim \mathcal{N}(0,\mathbf{I}),
\end{equation}
where $\bar{\alpha}_{\tau}$ is the cumulative noise schedule. 
The transformer receives the noisy future latent tokens $S_{t+\Delta t}^{\tau}$ and predicts the injected noise conditioned on the context latents, the 6-DoF relative action, the relative time, and the diffusion step:
\begin{equation}
    \hat{\epsilon}_{\theta}
    =
    f_{\theta}\left(
    S_{t+\Delta t}^{\tau}, \tau,
    S_{t-c+1:t}, e_a(a_{t\rightarrow t+\Delta t}), e_{\Delta}(\Delta t)
    \right),
\end{equation}
where $e_a(\cdot)$ embeds the relative probe motion and $e_{\Delta}(\cdot)$ embeds the temporal offset.
As for the 6-DoF probe action, a cosine embedding is used to represent the translation and rotation on different scales. The action is normalized by dataset-specific translation and rotation scales to obtain a more consistent representation across different neck scanning targets. A multi-scale cosine embedding is then applied to the normalized action:

\begin{equation}
  \hat{a} = \left[\frac{a^{1:3}}{s_{trans}}, \frac{a^{4:6}}{s_{rot}}\right]
\end{equation}
\begin{equation}
  e_a(\hat{a}) = \left[
    \sin\left(\frac{\pi \hat{a}}{2^l}\right),
    \cos\left(\frac{\pi \hat{a}}{2^l}\right)
  \right]_{l=0}^{L-1}
\end{equation}

The denoising objective of the CDiT is to minimize the mean squared error between the predicted noise and the true noise:
\begin{equation}
    \mathcal{L}_{\mathrm{wm}}
    =
    \mathbb{E}_{S,\epsilon,\tau}
    \left[
    \left\|
    \epsilon -
    \hat{\epsilon}_{\theta}
    \right\|_2^2
    \right].
\end{equation}

At inference stage, the model starts from Gaussian latent noise and iteratively denoises it under the specified context-action-time condition. 
The generated latent is then decoded into an ultrasound frame using the VAE decoder $dec$:
\begin{equation}
    \hat{I}_{t+\Delta t}
    =
    dec\left(\hat{S}_{t+\Delta t}\right),
\end{equation}
which provides the predicted visual consequence of the candidate probe motion. 
By modeling future observations as a conditional distribution instead of a deterministic regression target, the latent diffusion world model can better handle the ambiguity and appearance uncertainty caused by tissue deformation, probe pressure variation, and acoustic artifacts.

\subsection{Goal-conditioned action predictor} \label{sec:III-C}

\begin{figure*}[h]
  \centering
  \includegraphics[width=0.85\linewidth]{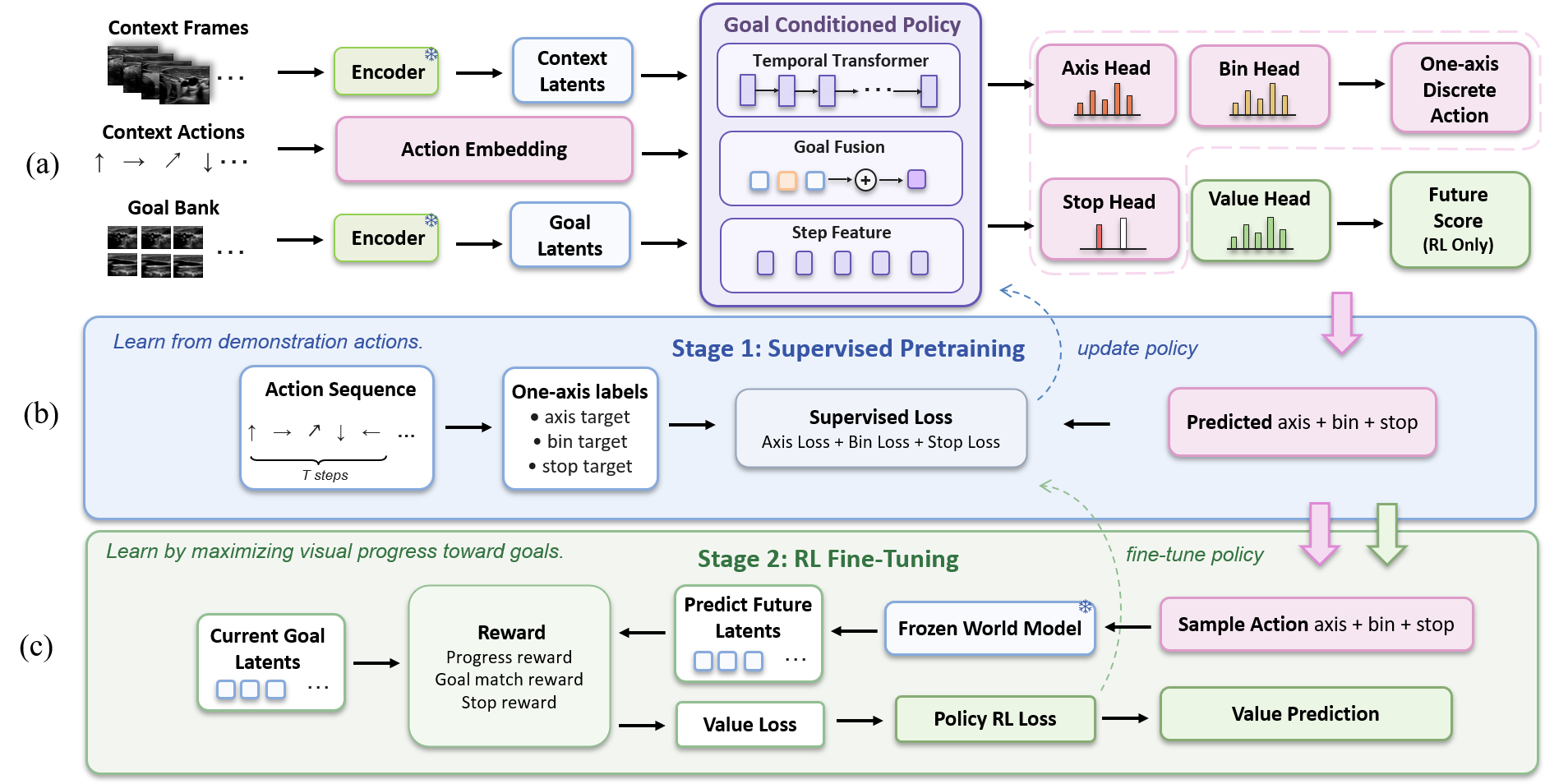}
  \caption{Goal-conditioned action predictor and training pipeline. Context frames, previous actions, and goal-bank latents are encoded into a temporal transformer policy. \textbf{Stage 1} trains the axis, bin, and stop heads using supervised demonstration labels. \textbf{Stage 2} finetunes the policy with rewards from the frozen world model.}
  \label{fig:predictor}
  \vspace{-6mm}
\end{figure*}

\textbf{Goal bank}
The action predictor is designed to move the probe towards an ordered sequence of target standard planes rather than to imitate a single target motion. The target sequence and visual criteria for each goal were specified by a sonographer. For the carotid task \cite{landwehr2001ultrasound}, 
the ordered goals correspond to reaching the carotid bifurcation and subsequently acquiring a longitudinal vessel view.
\label{part:thyroid_goals}
For the thyroid task, the goals correspond to centering the thyroid, moving towards head until it disappears, returning to the largest transverse section, and rotating to the longitudinal view.
Let the scanning protocol contain a goal bank
\begin{equation}
    G=\{G_1,G_2,\ldots,G_K\}
\end{equation},
where each $G_k$ denotes one target standard plane that sonographer has defined. 
Because a standard ultrasound plane can vary across patients, probe pressure, and small pose changes, each goal is represented by multiple latent samples instead of a single prototype:
\begin{equation}
    G_k=\{z_k^1,z_k^2,\ldots,z_k^{N_k}\}, \quad
    z_k^j=enc(I_k^j).
\end{equation},
where $I_k^j$ is the $j$-th annotated image for the $k$-th goal plane, and the encoder $enc(\cdot)$ is the same frozen VAE encoder used by the world model. 
During training, the goal bank is constructed from annotated key frames in the training dataset. 
At inference time, the robot follows the fixed protocol $G_1\rightarrow G_2\rightarrow\cdots\rightarrow G_K$ based on the current context ultrasound images and actions.

As shown in Fig. \ref{fig:predictor}a, the actor takes the recent latent context $S_{t-c+1:t}$, the corresponding previous relative actions $\{a_{t-c\rightarrow t-c+1},\ldots,a_{t-1\rightarrow t}\}$, and the goal bank $G$ as input.
The context frames are projected into latent tokens and combined with the previous action tokens. 
A temporal transformer then summarizes the recent scanning history into a context feature. 
For each goal plane, the model projects all latent samples in $G_k$ and pools them with learned sample weights, so the goal condition remains tied to several valid target appearances rather than a single averaged image.

\textbf{Axis, bin and stop head}
For a prediction horizon of $H$, the policy predicts a sequence of goal-conditioned corrective actions. At each output step $h\in\{0,1,\ldots,H-1\}$, the corresponding goal plane is selected according to the target-plane schedule along the demonstration trajectory.
The latent representation of the selected goal plane is then fused with the encoded context feature to obtain a goal-conditioned step feature, denoted as $\mathrm{step}_{t,h}$. Based on $\mathrm{step}_{t,h}$, the actor produces three outputs: an axis prediction $\hat{y}^{\mathrm{axis}}_{t,h}$, a motion-bin prediction $\hat{y}^{\mathrm{bin}}_{t,h}$, and a stop prediction $\hat{y}^{\mathrm{stop}}_{t,h}$. 
The axis head predicts the dominant motion dimension to be adjusted at the current step, the bin head predicts the discrete motion magnitude along the selected axis, and 
the stop head predicts whether the current observation is consistent with the sonographer-defined goal representation.

Instead of directly regressing continuous multi-axis probe motions, we formulate action prediction as a single-axis discrete classification problem.
This design is motivated by the closed-loop nature of robotic ultrasound navigation, where the probe can be progressively corrected based on updated image feedback. For each training action, we first identify the dominant axis as the dimension with the largest normalized action magnitude:
\begin{equation}
    y^{\mathrm{axis}}_{t,h}
    =
    \arg\max_{d}
    \frac{|a^{d}_{t,h}|}{s_d},
\end{equation}
where $a^{d}_{t,h}$ denotes the continuous action value along dimension
$d$, and $s_d$ is a dimension-specific normalization factor estimated from the training set. The motion value on the selected axis is then converted into one of five ordered bins: large negative motion, small negative motion, no motion, small positive motion, and large positive motion. 
The bin boundaries are calibrated from the training set to provide the small and heterogeneous motion ranges of ultrasound probe movements across different axes. This single-axis discretization reduces the action space, yields bounded and interpretable motion primitives, and allows the world model to provide more direct and stable reward signals for policy learning.

The stop target is defined according to the target-plane offsets rather than simply marking the final step of a trajectory. Given the set of target-plane arrival offsets $\{o_1,o_2,\ldots,o_K\}$, the stop label is assigned as
\begin{equation}
    y^{\mathrm{stop}}_{t,h}
    =
    \mathbf{1}\left[
    t+h\in\{o_1,o_2,\ldots,o_K\}
    \right].
\end{equation}

Therefore, the stop head learns to recognize the arrival of each intermediate target plane. When a stop decision is triggered for the current active goal, the policy advances to the next target in the bank.

As shown in Fig.~\ref{fig:predictor}b and Fig.~\ref{fig:predictor}c, the action predictor is trained with a supervised action-bin loss and a world-model perceptual reward.

\textbf{The first training stage} uses supervised imitation from expert trajectories. 
The main loss is a weighted multi-head classification objective:
\begin{equation}
    \mathcal{L}_{sup}
    =
    \lambda_{axis}\mathcal{L}_{axis}
    +
    \lambda_{bin}\mathcal{L}_{bin}
    +
    \lambda_{stop}\mathcal{L}_{stop}.
\end{equation}
$\mathcal{L}_{axis}$ is the cross entropy loss for the active axis, $\mathcal{L}_{bin}$ is the per-axis cross entropy loss for the five-bin action labels on the active dimensions, and $\mathcal{L}_{stop}$ is the cross entropy loss for the stop decision. 
The axis and bin losses are applied to non-stop steps, while the stop loss is applied over the whole horizon. 

\textbf{The second training stage} is to fine-tune the same policy with the frozen ultrasound world model described above using actor-critic RL. 
For a selected current goal $G_k$, the actor predicts an action axis, a motion bin and a stop decision from the current context.
If the stop decision is not triggered, the predicted bins are converted back into a valid relative probe motion $a_{t \rightarrow t+1}$. The frozen world model then rolls out the next latent ultrasound state:
\begin{equation}
    \hat{S}_{t+1}
    \sim
    W\left(
    S_{t-c+1:t}, a_{t \rightarrow t+1}, \Delta t=1
    \right).
\end{equation}

To measure progress toward the current target plane, we compute the distance between the predicted latent state and the target-plane distribution represented by the goal bank. Since each target plane $G_k$ contains multiple latent samples, 
the current distance and predicted next distance are defined as $d_t^k = \min_j \frac{1}{D}\left\|S_t-z_k^j\right\|_2$ and $\hat{d}_{t+1}^k = \min_j \frac{1}{D}\left\|\hat{S}_{t+1}-z_k^j\right\|_2$, where $D$ is the number of latent elements. The reward is then defined as:
\begin{equation}
\begin{aligned}
    R_t =
    &\ \alpha\left(d_t^k-\hat{d}_{t+1}^k\right)
    -\beta \hat{d}_{t+1}^k
    + R_{\mathrm{stop}} \\
    &-\lambda_a\left\|a_{t\rightarrow t+1}\right\|_1
    -\lambda_s\left\|a_{t\rightarrow t+1}-a_{t-1\rightarrow t}\right\|_1 .
\end{aligned}
\end{equation}

The first term rewards the relative progress toward the active target plane, while the second term encourages the predicted state to keep close to the target distribution. The stop reward $R_{stop}$ evaluates whether the policy stops at the correct target plane:
\begin{equation} 
  R_{\mathrm{stop}} = 
  \begin{cases}
     r_{\mathrm{succ}}, & \text{if } \hat{y}^{\mathrm{stop}}_t=1 \text{ and } d_t^k < \tau_k,\\
     -r_{\mathrm{prem}}, & \text{if } \hat{y}^{\mathrm{stop}}_t=1 \text{ and } d_t^k \geq \tau_k,\\ 
     0, & \text{otherwise}, 
  \end{cases} 
\end{equation}
where $\tau_k$ is a target-specific reach threshold calibrated from held-out examples of the $k$ th standard plane. When a successful stop is triggered, the active goal is advanced from $G_k$ to $G_{k+1}$. The last two terms regularize the motion by penalizing unnecessarily large actions and abrupt action changes, controlled by $\lambda_a$ and $\lambda_s$, respectively.

During this stage, the reward signal is used to update both the action policy and the value head. To keep the value estimate consistent with the actor input and the motion-smoothness reward, we define the policy state as
\begin{equation}
    x_t=\left(S_{t-c+1:t}, A_t^{\mathrm{hist}}, G_k\right),
    \quad
    A_t^{\mathrm{hist}}=\{a_{t-c\rightarrow t-c+1},\ldots,a_{t-1\rightarrow t}\}.
\end{equation}
After applying the sampled action, the next state is formed by appending $a_{t\rightarrow t+1}$ to the action history and using the predicted next latent $\hat{S}_{t+1}$ as the newest visual state. If the stop decision succeeds, the active goal in the next state is advanced from $G_k$ to $G_{k+1}$. The value head estimates the expected future return $V_{\phi}(x_t)$ under the current active goal and serves as a baseline for computing the advantage: 
\begin{equation} 
  \mathrm{Advantage}_t = R_t + \gamma V_{\phi}(x_{t+1}) - V_{\phi}(x_t),
\end{equation}
The advantage is then used as the policy learning signal for the categorical axis, bin, and stop heads.

A positive advantage increases the probability of the sampled axis, bin, and stop decisions, whereas a negative advantage suppresses them. The value head is trained to regress the bootstrapped return: 
\begin{equation} 
  \mathcal{L}_{\mathrm{value}} = \left\| V_{\phi}(x_t) - \left(R_t+\gamma V_{\phi}(x_{t+1})\right) \right\|_2^2 . 
\end{equation} 
Thus, the reward provides the learning signal from the frozen world model, while the value head learns to estimate the long-term expected progress toward the current protocol goal.

Both stages use AdamW with weight decay $10^{-4}$. Stage 1 is trained for $50$ epochs with learning rate $10^{-5}$ and an effective batch size of $48$. Stage 2 is trained for another $50$ epochs with learning rate $3\times10^{-6}$ and an effective batch size of $24$.

\begin{figure*}[!t]
  \centering
    \includegraphics[width=0.9\linewidth]{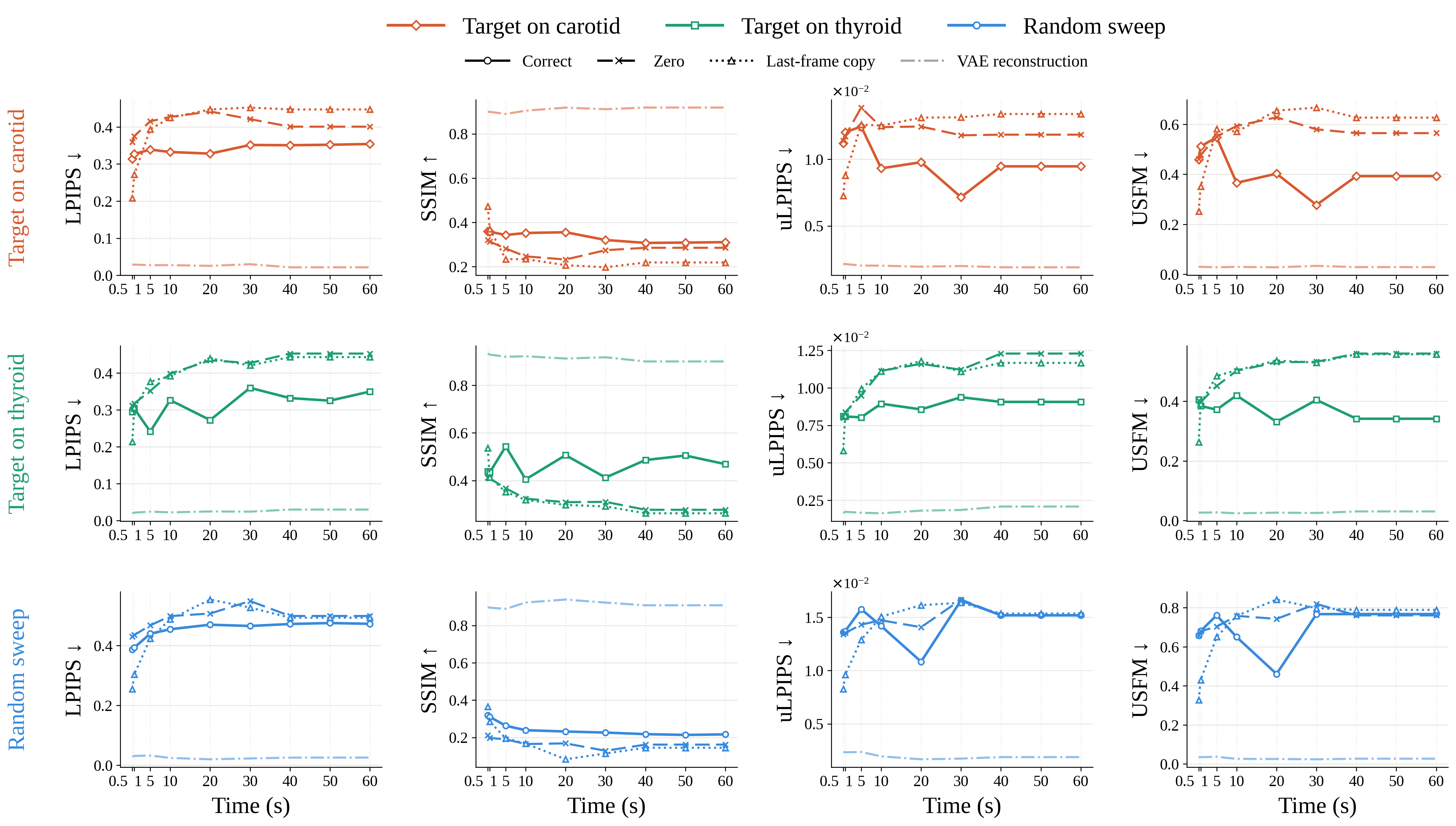}
    \caption{\revision{
  Evaluation of the action-conditioned ultrasound world model over time horizons across trajectory types. Mean LPIPS, SSIM, uLPIPS and USFM between predicted future frames and ground truth frames at different time offsets from the initial observation.
  }}
    \label{fig:wm_quality_horizon}
\vspace{-6mm}
\end{figure*}

\section{Experiment}\label{sec:exp}

\subsection{Experimental setup}

We evaluate the proposed framework from two complementary aspects: 
\begin{itemize}
  \item[(1)] whether the latent diffusion world model can serve as an action-conditioned simulator of ultrasound appearance,
  \item[(2)] whether the learned predictor can estimate useful probe actions from recent ultrasound observations. 
\end{itemize}
All experiments are conducted on a self-collected neck ultrasound trajectory dataset from 20 people. Each sample contains an ultrasound frame sequence and synchronized 6-DoF probe poses. 
We split the dataset by individual rather than frames to avoid temporally correlated samples appearing in both training and testing. 

The dataset contains three trajectory groups. The first two groups are target directed scans around the carotid artery and thyroid gland. The third group contains random sweeps for less constrained neck exploration. This design allows us to evaluate both structured target seeking motion and broader local exploration. Table \ref{tab:dataset_stats} summarizes the dataset statistics. 



\begin{table}[htbp]
    \centering
    \caption{Statistics of the self-collected neck ultrasound trajectory dataset.}
    \label{tab:dataset_stats}
    \resizebox{\columnwidth}{!}{%
    \begin{tabular}{lccccc}
    \hline
    Subset & Traj & Train & Test & Frames  & Avg Horizon (s) \\ \hline
    Carotid & 184 & 147 & 37& 196,358  & 14.73 \\
    Thyroid & 150& 120 & 30   & 518,161& 47.41 \\
    Random sweep & 17 & 13 & 4 & 135,353 & 104.79 \\ \hline
    \end{tabular}%
    }
    \vspace{-6mm}
\end{table}

\subsection{Action-conditioned ultrasound world model evaluation}


\revision{This experiment evaluates whether the world model captures the visual consequences of probe motion. Given recent context frames, a relative 6-DoF probe action, and a temporal offset, the model predicts the future ultrasound frame. We evaluate prediction quality across different time horizons using image similarity metrics and controlled baselines. We further assess action sensitivity through counterfactual-action discrimination and anatomical accuracy
through organ segmentation Dice scores.}




\textbf{Prediction quality versus time}
\revision{Fig.~\mbox{\ref{fig:wm_quality_horizon}} compares prediction quality across carotid, thyroid and random-sweep trajectories over different time horizons. We evaluate LPIPS, SSIM, and two ultrasound-specific metrics, uLPIPS and USFM ~\mbox{\cite{huang2026defining}}, using correct actions (line Correct), zero-action inputs (line Zero) and a last-frame-copy baseline. VAE reconstruction is also included to separate reconstruction error from world-model prediction error.}

\revision{Correct action predictions generally outperform the zero action and last-frame-copy baselines, especially for target-directed trajectories. VAE reconstruction achieves substantially better image similarity. Random swpeeps show larger errors and less consistent improvements at longer horizons. These results suggest that the world model captures action-dependent ultrasound dynamics more reliably in structured neck scanning trajectories.}

\textbf{\revision{Action discrimination.}}
\revision{To further evaluate action sensitivity, we conduct a counterfactual-action experiment, as shown in Table~\mbox{\ref{tab:wrongaction}}. For each sample, we compare predictions using the recorded action with those using three incorrect actions from the same trajectory, preferentially selected within 20\% of the recorded action magnitude. 
The visual context and temporal offset remain unchanged. We report differences in LPIPS, SSIM, uLPIPS, and USFM, along with identification accuracy, defined as the proportion of cases in which human experts correctly identify the prediction generated using the ground-truth action.}

\revision{The correct actions generally produce predictions closer to the ground-truth future frames. At 20 seconds, identification accuracy reaches 91.7\% for carotid and 100\% for thyroid. However, random sweeps show weaker metrics at longer horizons despite an ACC of 100\%. These results indicate that the world model is more sensitive to probe actions in structured, target-directed trajectories. 
}

\begin{table}[!t]
  \centering
  \caption{\revision{Counterfactual-action discrimination across prediction horizons. }}
  \label{tab:wrongaction}
  \resizebox{\columnwidth}{!}{%
  \begin{tabular}{
      p{0.8cm}
      p{0.8cm}
      ccccc
    }
    \hline
    \multirow{2}{*}{Target}
    & \multirow{2}{*}{Time (s)}
    & \multirow{2}{*}{$\Delta$LPIPS $\downarrow$}
    & \multirow{2}{*}{$\Delta$SSIM $\uparrow$}
    & $\Delta$uLPIPS $\downarrow$
    & \multirow{2}{*}{$\Delta$USFM $\downarrow$}
    & Identification \\
    & & & & $(\times10^{-2})$ & & ACC $\uparrow$ \\
    \hline
  \multirow{6}{*}{Carotid} & 0.5 & -0.0027 & +0.0007 & -0.0010 & -0.0010 & 30.0\% \\ 
   & 1 & -0.0058 & +0.0027 & -0.0039 & -0.0079 & 52.8\% \\ 
   & 5 & -0.0185 & +0.0225 & -0.0350 & -0.0366 & 60.0\% \\ 
   & 10 & -0.0450 & +0.0289 & -0.1293 & -0.0658 & 70.0\% \\ 
   & 20 & -0.0845 & +0.0686 & -0.2286 & -0.1206 & 91.7\% \\ 
   & 30 & \textbf{-0.1011} & \textbf{+0.0732} & \textbf{-0.2743} & \textbf{-0.1514} & \textbf{100.0}\% \\ \hline
  \multirow{6}{*}{Thyroid} & 0.5 & +0.0051 & -0.0029 & +0.0029 & +0.0027 & 41.7\% \\
   & 1 & -0.0023 & -0.0040 & -0.0120 & +0.0023 & 45.0\% \\
   & 5 & -0.0411 & +0.0433 & -0.0987 & -0.0625 & 66.7\% \\
   & 10 & \textbf{-0.0882} & +0.0677 & -0.2233 & -0.0992 & 88.9\% \\ 
   & 20 & -0.0865 & +0.0725 & \textbf{-0.2313} & \textbf{-0.1077} & \textbf{100.0\%} \\
   & 30 & -0.0733 & \textbf{+0.0857} & -0.2085 & -0.1007 & 83.3\% \\ \hline
  \multirow{6}{*}{Random} & 0.5 & -0.0055 & -0.0004 & -0.0197 & -0.0029 & 30.6\% \\ 
   & 1 & -0.0086 & +0.0188 & -0.0022 & -0.0162 & 43.1\% \\
   & 5 & -0.0134 & +0.0180 & -0.0706 & -0.0257 & 33.3\% \\
   & 10 & -0.0147 & +0.0144 & -0.0011 & -0.0180 & 50.0\% \\
   & 20 & \textbf{-0.0510} & \textbf{+0.0592} & \textbf{-0.1009} & \textbf{-0.0682} & 90.0\% \\
   & 30 & +0.0077 & -0.0224 & -0.0082 & +0.0203 & \textbf{100.0\%}\\ \hline
  \end{tabular}%
}
\vspace{-7mm}
\end{table}

\begin{figure}[!t]
  \centering
  \includegraphics[width=\linewidth]{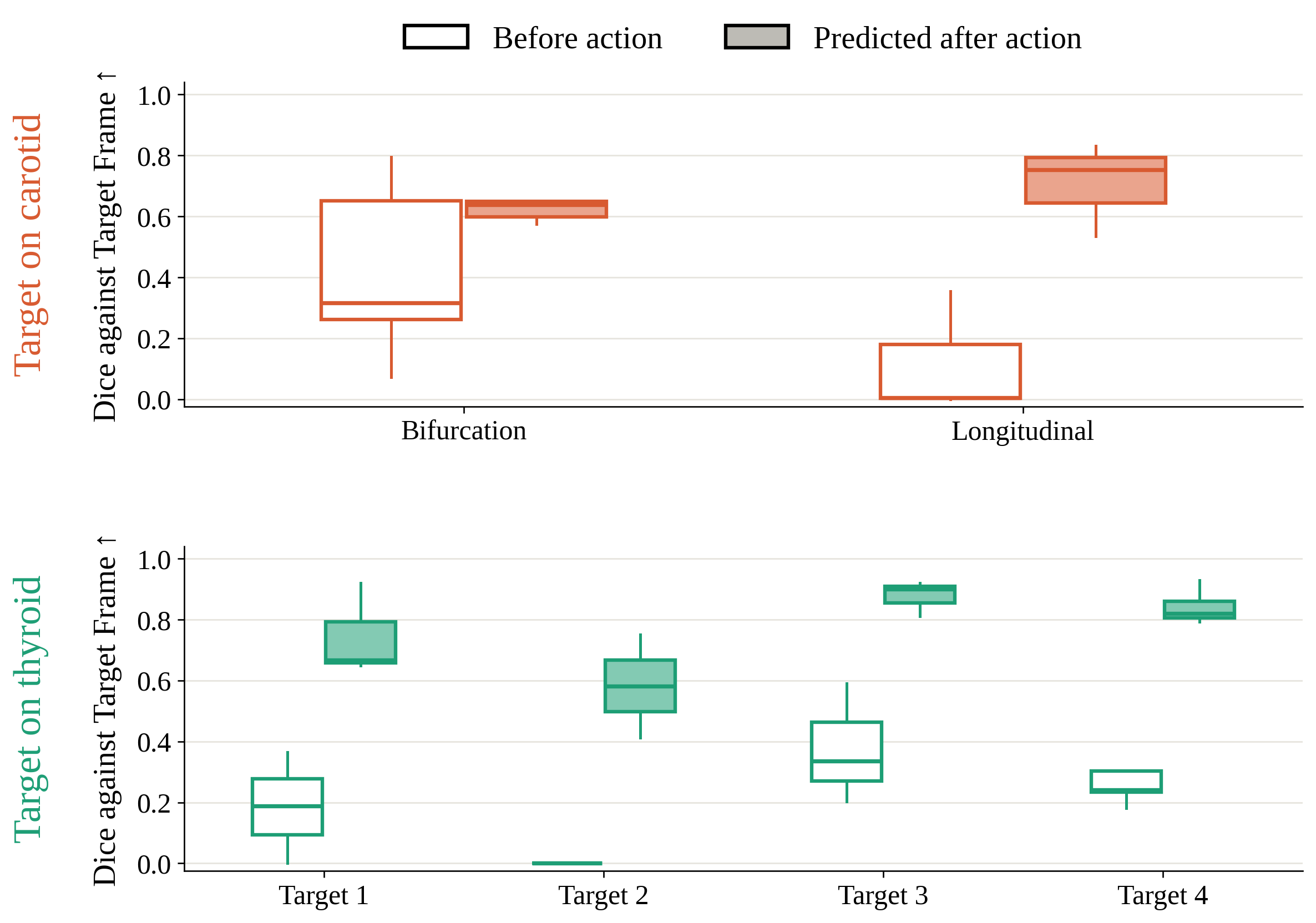}
  \caption{\revision{Task-specific anatomical evaluation. Boxplots show Dice distributions for carotid goals (top) and thyroid goals (bottom). Higher scores indicate better target-frame alignment.}}
  \label{fig:task_dice}
\end{figure}

\revision{\textbf{Task-specific anatomical evaluation.}}
\revision{
To further evaluate anatomical prediction accuracy, we compare organ segmentation Dice scores before and after the predicted probe motion, as shown in
Fig.~\mbox{\ref{fig:task_dice}}. The world model predicts the future frame using the recorded relative 6-DoF motion toward the target plane. Segment Dice scores using UltraSAM3~\mbox{\cite{xu2026ultrasam3}} are computed between the target frame and the current or predicted frame. The predicted frames achieve higher Dice scores and generally lower variability across carotid and thyroid goals. These results suggest that the world model captures anatomically meaningful changes under target-directed probe motions.
}

\vspace{-6mm}
\subsection{Goal-conditioned action predictor evaluation}

\begin{figure}[!t]
  \centering
  \includegraphics[width=\linewidth]{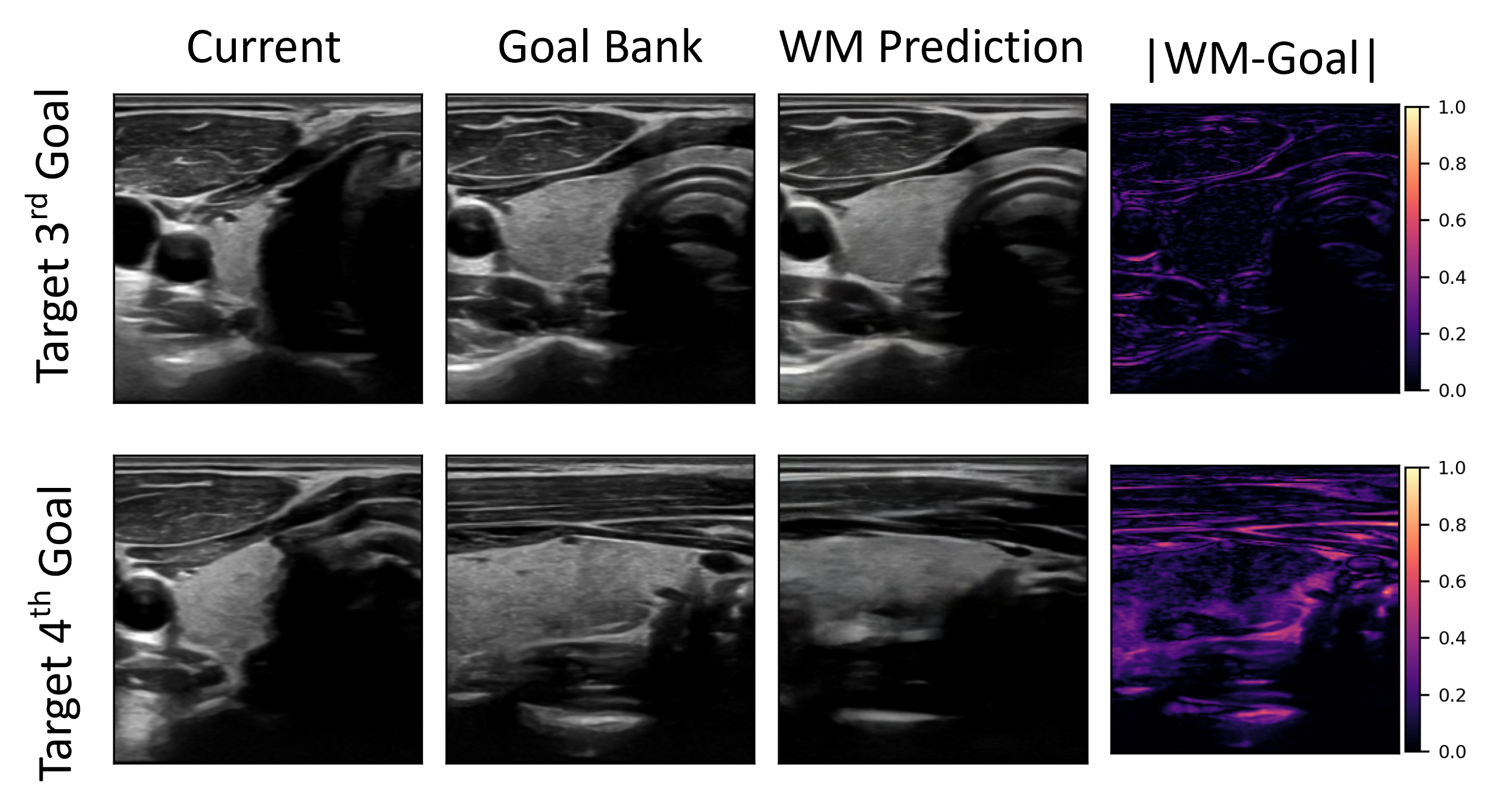}
  \caption{\revision{Qualitative thyroid trajectory prediction from the goal conditioned action predictor. $|\text{WM}-\text{Goal}|$ denotes the per pixel absolute error, darker is better.}}
  \label{fig:predictor_sample}
  \vspace{-8mm}
\end{figure}

\begin{table*}[!h]
    \centering
    \caption{\revision{Prediction accuracy for ordered goal plane guidance.}}
    \label{tab:predictor}
    \resizebox{0.95\textwidth}{!}{%
    \begin{tabular}{llcccccccccc}
    \hline
    Target & Method & S1 Goal & S2 Goal & Axis $\uparrow$ & $x$ bin $\uparrow$ & $y$ bin $\uparrow$ & $z$ bin $\uparrow$ & $r_x$ bin $\uparrow$ & $r_y$ bin $\uparrow$ & $r_z$ bin $\uparrow$ & Stop F1 $\uparrow$ \\
    \hline
    \multirow{5}{*}{Carotid} & Supervised only & $\times$ & -- & 0.231 & 0.129 & 0.192 & 0.438 & 0.249 & 0.077 & 0.183 & -- \\
    & Supervised only & $\checkmark$ & -- & 0.786 & 0.134 & 0.191 & \textbf{0.970} & 0.888 & 0.492 & 0.246 & 0.835 \\
     & + RL finetuning & $\times$ & $\times$ & 0.214 & 0.127 & 0.191 & 0.433 & 0.201 & 0.079 & 0.217 & -- \\
    & + RL finetuning & $\checkmark$ & $\times$ & \textbf{0.799} & \textbf{0.764} & \textbf{0.685} & 0.851 & 0.977 & 0.951 & 0.781 & 0.846  \\
     & + RL finetuning & $\checkmark$ & $\checkmark$ & \textbf{0.799} & 0.763 & 0.683 & 0.851 & \textbf{0.978} & \textbf{0.956} & \textbf{0.791} & \textbf{0.883}  \\
    \hline
    \multirow{5}{*}{Thyroid} & Supervised only & $\times$ & -- & 0.256 & 0.282 & 0.304 & 0.214 & 0.304 & 0.138 & 0.174 & -- \\
    & Supervised only  & $\checkmark$ & -- & \textbf{0.625} & 0.482 & \textbf{0.941} & \textbf{0.684} & \textbf{0.897} & \textbf{0.813} & \textbf{0.692} &  0.704 \\
    & + RL finetuning  & $\times$ & $\times$ & 0.248 & 0.304 & 0.317 & 0.199 & 0.290 & 0.147 & 0.174 & -- \\
    & + RL finetuning  & $\checkmark$ & $\times$ & 0.528 & 0.482 & 0.588 & \textbf{0.684} & 0.862 & 0.625 & 0.200 & 0.648\\
    & + RL finetuning  & $\checkmark$ & $\checkmark$ & \textbf{0.625} & \textbf{0.667} & \textbf{0.941} & \textbf{0.684} & \textbf{0.897} & \textbf{0.813} & \textbf{0.692} & \textbf{0.762} \\
    \hline
    \end{tabular}%
}
\vspace{-5mm}
\end{table*}

This section evaluates whether the proposed goal-conditioned action predictor can estimate corrective probe motions toward the ordered target planes. \revision{Fig.~\mbox{\ref{fig:predictor_sample}} shows qualitative predictions across two out of four thyroid goals. }
Each row presents the current frame, the target goal plane image, the future frame generated by the world model conditioned on the predicted action, and the absolute difference map between the world-model prediction and the goal image. 
From target $1^{st}$ to $4^{th}$, corresponding to four stages described \hyperref[part:thyroid_goals]{in the goal-bank section}.
The predicted future frames move the visible anatomy toward the selected goal planes in these examples, and the error maps remain low over large image regions. This qualitative result suggests that the predicted actions produce visually meaningful goal directed changes under the learned ultrasound dynamics.

\revision{
Table~\mbox{\ref{tab:predictor}} evaluates the action predictor using axis accuracy, per-axis motion-bin accuracy, and Stop F1.
We compare supervised-only policies with world-model-guided fine-tuning under different goal-conditioning settings.
The S1 Goal and S2 Goal columns indicate whether goal representations are used in each training stage.
Stop F1 is not reported for policies trained without goal conditioning from Stage 1, since goal-specific stopping decisions are undefined.
}

\revision{
Goal conditioning substantially improves action prediction.
In Stage 1, axis accuracy increases from 0.231 to 0.786 for carotid and from 0.256 to 0.625 for thyroid guidance.
This advantage remains after Stage 2.
The fully goal-conditioned policies achieve axis accuracies of 0.799 and 0.625, compared with 0.214 and 0.248 for
the independently trained no-goal baselines.
These results indicate that goal information is essential for selecting appropriate corrective motions.
}

\revision{
World-model-guided fine-tuning addresses different weaknesses of supervised learning across the two tasks.
For carotid guidance, the supervised policy already predicts the active axis accurately, but performs poorly on several motion bins. Fine-tuning substantially improves the x, y, rx, ry, and rz bin accuracies, while Stop F1 increases from 0.835 to 0.883.
However, z-bin accuracy decreases from 0.970 to 0.851. For thyroid guidance, supervised learning already achieves high accuracy on several motion bins, leaving less room for improvement. Fine-tuning mainly improves the weaker x-bin accuracy from 0.482 to 0.667 and Stop F1 from 0.704 to 0.762.
These results suggest that world-model-guided fine-tuning can complement supervised learning by refining task-dependent weaknesses, although the improvements are not uniform across motion dimensions.
}

\revision{
The role of goal conditioning during Stage 2 also differs between the two tasks.
For carotid guidance, removing the goal input during fine-tuning has little effect on axis and bin accuracies, but reduces Stop F1 from 0.883 to 0.846. 
This suggests that goal-conditioned supervised training already provides sufficient information for most motion predictions in the two-goal carotid protocol.
For thyroid guidance, retaining the goal input improves axis accuracy from 0.528 to 0.625 and Stop F1 from 0.648 to 0.762, together with several motion-bin accuracies.
The four-stage thyroid protocol involves direction reversals and visually similar intermediate views, making explicit goal information more important during fine-tuning.
Overall, goal conditioning helps resolve ambiguity between protocol stages, while world-model-guided fine-tuning provides task-specific refinement of motion and stop predictions.
}

\subsection{In the real-world ultrasound scenario}
\vspace{-3mm}
\begin{figure}[htbp]
  \centering
  \includegraphics[width=0.6\linewidth]{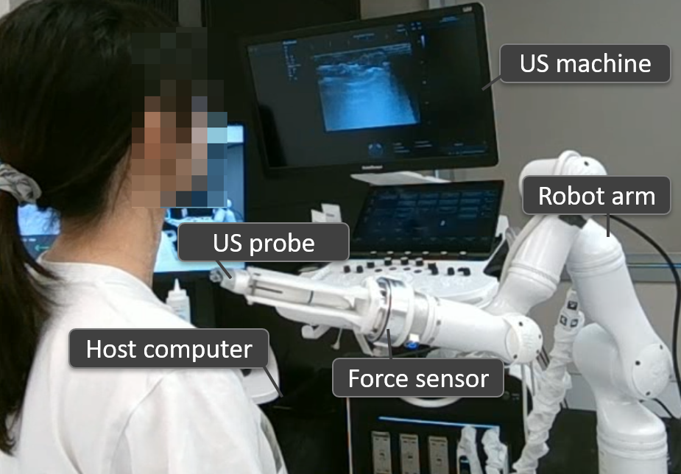}
  \caption{The physical setup.}
  \label{fig:realworld}
  \vspace{-5mm}
\end{figure}

In this section, we evaluate whether the learned predictor can be deployed as a closed-loop guidance module on a real robotic ultrasound platform.
Fig.~\ref{fig:realworld} shows the robotic ultrasound system, which consists of an ultrasound imaging system (S80, SonoScape, China) with a linear probe mounted on a 6-axis collaborative robot arm (RM65-B, RealMan, China).
A 6-axis force/torque sensor (PhotonR75, Haptron, China) is integrated into the robot end-effector to measure the probe to skin interaction during scanning.

In each trial, the robot starts from an off-target neck view and receives live ultrasound frames.
The goal-conditioned predictor, with or without WM fine-tuning, estimates the next discrete corrective motion and executed in a closed loop.
The trial ends after all goal planes are reached or the maximum step number is exceeded.
For subject safety, only $x$, $y$, $r_x$, and $r_z$ are sent as position commands, while $z$ and $r_y$ directions are controlled by a low level force control module following \cite{cao2023ultra}.
We evaluate carotid and thyroid scanning tasks from nearby off target initial views in Table \ref{tab:realworld_results}. For Goal $k$, the success rate is the proportion of trials that the robot reaches the $k$th goal plane without manual intervention and within $100$ seconds.


\begin{table}[!t]
    \centering
    \caption{\revision{Real-world scanning results (20 trials per task). Steps and time are averaged over successful trials.}}
    \label{tab:realworld_results}
    \resizebox{\columnwidth}{!}{%
    \begin{tabular}{lllccc}
    \hline
    Task & Goal & Method & Success Rate & Avg Steps & Avg Time (s) \\ \hline
    \multirow{4}{*}{Carotid}
    & \multirow{2}{*}{Goal 1}
    & Supervised only & 0.90 & 722 & 24.1 \\
    & & + WM fine-tuning & 0.90 & 697 & 23.3 \\
    \cline{2-6}
    & \multirow{2}{*}{Goal 2}
    & Supervised only & 0.30 & 1725 & 57.6 \\
    & & + WM fine-tuning & 0.70 & 1290 & 43.1 \\ \hline

    \multirow{8}{*}{Thyroid}
    & \multirow{2}{*}{Goal 1}
    & Supervised only & 0.80 & 485 & 16.2 \\
    & & + WM fine-tuning & 0.85 & 445 & 14.8 \\
    \cline{2-6}
    & \multirow{2}{*}{Goal 2}
    & Supervised only & 0.70 & 1501 & 50.1 \\
    & & + WM fine-tuning & 0.80 & 701 & 23.5 \\
    \cline{2-6}
    & \multirow{2}{*}{Goal 3}
    & Supervised only & 0.20 & 2491 & 83.2 \\
    & & + WM fine-tuning & 0.75 & 841 & 28.2\\
    \cline{2-6}
    & \multirow{2}{*}{Goal 4}
    & Supervised only & 0.00 & -- & -- \\
    & & + WM fine-tuning & 0.65 & 1046 & 34.9 \\ \hline
    \end{tabular}%
    }
    \vspace{-5mm}
\end{table}


\revision{
WM-guided fine-tuning improves closed-loop navigation over supervised learning, increasing the final success rate from 0.30 to 0.70 for carotid and from 0.00 to 0.65 for thyroid, while reducing execution time at thyroid Goal 3 from 83.2 s to 28.2 s. As shown in Table.\mbox{\ref{tab:predictor}}, these gains correspond to improved multi-axis motion-bin predictions for carotid and higher $x$-axis accuracy and stop F1 for thyroid. Improved motion correction and stopping decisions may enhance goal-reaching success and reduce unnecessary movements, particularly during later stages of multi-goal scanning.

Nevertheless, the single-axis action formulation remains
a limitation. Some target views require coordinated
corrections along multiple axes, while the current policy
adjusts only one axis at each step. This can leave residual
pose errors unresolved and eventually cause navigation
failure.}

\section{Discussion \& Future work}
The results suggest that the action-conditioned world model provides a useful training signal for robotic ultrasound guidance, but its scope remains limited. The model learns from ultrasound image transitions, relative probe motion, and temporal offsets, making it suitable for local, target-directed neck scanning. However, it does not explicitly distinguish probe-induced changes from physiological motion or contact variation.
\revision{
The results support its use in neck ultrasound guidance, but generalization beyond this setting remains unverified.
}

Respiration, swallowing, and organ motion can alter tissue position, probe contact, and the acoustic window independently of robot actions. These effects are especially important for abdominal and other deformable anatomies.
\revision{
Future work could expand the dataset to more diverse anatomies and incorporate respiratory and temporal motion cues to better distinguish robot-induced changes from anatomical motion.
}

Another limitation is that force control is used only at the robot control level and is not used as an input modality for either the world model or the action predictor. Consequently, the framework cannot directly model how probe pressure affects tissue deformation and image quality. Incorporating force-torque measurements as an additional modality may enable pressure-aware dynamics and support actions that jointly optimize goal progress, contact stability, and patient safety.

\section{Conclusion}
\revision{
This paper presented an action conditioned world model for goal plane guidance in robotic neck ultrasound.
}
The core idea is to learn visual action dynamics from neck ultrasound trajectories and use the frozen world model as an internal simulator for training a goal conditioned action predictor.
The latent diffusion world model predicts future ultrasound observations from context frames, relative probe motion, and temporal offset, while the action predictor uses goal bank latents to select ordered corrective actions.

Experiments on self collected neck ultrasound trajectories show that the learned world model preserves action dependent anatomical structure more reliably in target directed scans than in random sweeps, while the guided fine-tuning improves several translation and rotation action predictions. In the real world online evaluation, the system achieves success rates of $70.0\%$ for carotid guidance and $65.0\%$ for thyroid guidance.

These results demonstrate the potential of learned ultrasound dynamics as a training signal for robot probe navigation. Future work will incorporate force information to improve robustness across anatomies and scanning conditions.



\bibliographystyle{ieeetr}
\bibliography{bibliography}

\end{document}